\pdfoutput=1
\documentclass[11pt]{article}

\usepackage[]{EMNLP2023}

\usepackage{times}
\usepackage{latexsym}

\usepackage[T1]{fontenc}
\usepackage[utf8]{inputenc}

\usepackage{microtype}

\usepackage{inconsolata}

\usepackage{graphicx}
\usepackage{amsmath}
\usepackage{amsfonts,amssymb}
\usepackage{multirow}
\usepackage{booktabs}
\usepackage{hyperref}
\usepackage{float}
\usepackage{enumitem}
\usepackage{algorithmicx}
\usepackage{algpseudocode}
\usepackage{tabularx}
\usepackage{textgreek}

\newcommand{\modelname}{\textsc{CoUDA}}

\definecolor{Red}{RGB}{255,0,0}
\definecolor{Green}{RGB}{0,176,80}
\definecolor{Blue}{RGB}{48,184,232}
\definecolor{Purple}{RGB}{138,89,179}

\title{CoUDA: Coherence Evaluation via Unified Data Augmentation}

\author{ \
Dawei Zhu~\thanks{~~~Dawei Zhu and Wenhao Wu contribute equally to this paper. Prof. Sujian Li is the corresponding author.}~~$^\text{\texteta~\textdelta}$
\quad Wenhao Wu~\footnotemark[1]~~$^\text{\texteta~\textdelta}$
\quad Yifan Song~$^\text{\texteta~\textdelta}$
\quad Fangwei Zhu~$^\text{\texteta~\textdelta}$
\\{\bf
Ziqiang Cao~$^\text{\textpi}$
\quad Sujian Li~$^\text{\texteta~\textdelta~\textlambda}$}
\\
$^\text{\texteta}$~School of Computer Science, Peking University
\\
$^\text{\textdelta}$~National Key Laboratory for Multimedia Information Processing, Peking University \\
$^\text{\textpi}$~Institute of Artificial Intelligence, Soochow University \\
$^\text{\textlambda}$~Jiangsu Collaborative Innovation Center for Language Ability, Jiangsu Normal University
}

\begin{document}
\maketitle

\begin{abstract}

Coherence evaluation aims to assess the organization and structure of a discourse, which remains challenging even in the era of large language models. 
Due to the scarcity of annotated data, data augmentation is commonly used for training coherence evaluation models. 
However, previous augmentations for this task primarily rely on heuristic rules, lacking designing criteria as guidance.
In this paper, we take inspiration from linguistic theory of discourse structure, and propose a data augmentation framework named \modelname{}. \modelname{} breaks down discourse coherence into global and local aspects, and designs augmentation strategies for both aspects, respectively.
Especially for local coherence, we propose a novel generative strategy for constructing augmentation samples, which involves post-pretraining a generative model and applying two controlling mechanisms to control the difficulty of generated samples. 
During inference, \modelname{} also jointly evaluates both global and local aspects  to comprehensively assess the overall coherence of a discourse.
Extensive experiments in coherence evaluation show that, with only 233M parameters,  \modelname{} achieves state-of-the-art performance in both pointwise scoring and pairwise ranking tasks, even surpassing recent GPT-3.5 and GPT-4 based metrics. \footnote{~\url{https://github.com/dwzhu-pku/CoUDA}}

\end{abstract}

\section{Introduction}\label{intro}

\textit{Coherence} is a vital aspect of communication that evaluates the structure and organization of discourse~\citep{Halliday76a, groszAttentionIntentionsStructure1986}. Consequently, models capable of evaluating coherence of the given text are widely applicable in both discourse generation and assessment. While recent large language models show strong  performance in various tasks~\citep{brown2020language}, they have not presented superiority in coherence evaluation compared with the fine-tuning based models~\citep{fuGPTScoreEvaluateYou2023}. Considering both computational efficiency and evaluation performance a good evaluation metric should possess,  in this paper, we focus on modeling coherence via a fine-tuning based lightweight model.

\begin{figure}
    \centering
    % \phantom{\rule{6cm}{4cm}}
    \includegraphics[width=\linewidth]{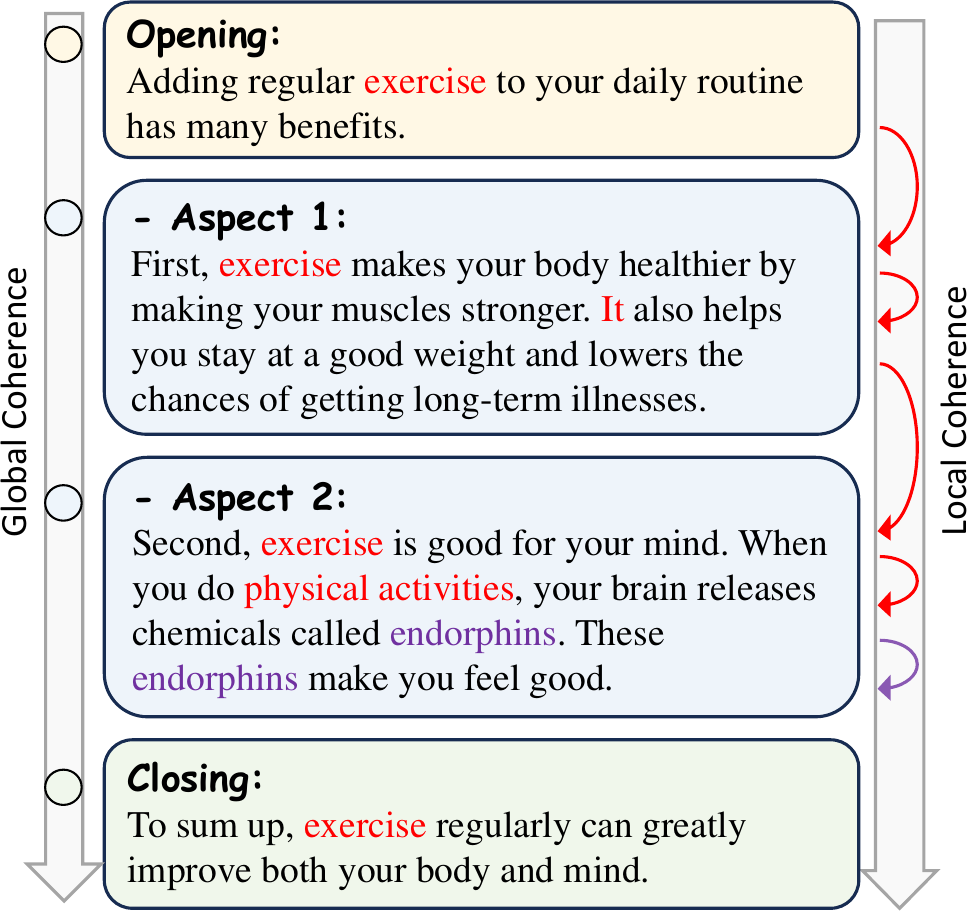}
    \caption{Example for global coherence and local coherence in a discourse.  Globally, the discourse is well-structured, with a opening sentence to introduce the argument, five sentences to give evidence from two aspects, and a closing sentence for conclusion. Locally, the focused items, which is denoted in \textcolor{Red}{Red} and \textcolor{Purple}{Purple}, transfers smoothly from sentence to sentence.
    }
    \label{fig:task_example}
\end{figure}

Due to the scarcity of human-annotated data, data augmentation techniques are commonly employed in training a coherence evaluation model
~\citep{liNeuralNetModels2017, jwalapuramRethinkingSelfSupervisionObjectives2022}.
As human-written discourses naturally possess coherence and can thus serve as positive samples, previous research has focused on constructing negative samples, primarily through rule-based methods such as swapping or shuffling sentences \citep{barzilayModelingLocalCoherence2008,shenEvaluatingDocumentCoherence2021,jwalapuramRethinkingSelfSupervisionObjectives2022}. 
However, as these methods are heuristically inspired without any design criteria as guidance, they suffer from weak correlation with human judgements~\citep{mohiuddinRethinkingCoherenceModeling2021}.
This brings up the research question: To effectively model coherence, can we find reasonable criterium as guidance to design augmentation strategies?

According to \citet{groszAttentionIntentionsStructure1986}, discourse coherence is mainly determined by two aspects: the organization of discourse segments~(i.e. \textit{global coherence}), and the transition of attention or focused items~(i.e. \textit{local coherence}). Examples for these two aspects of coherence are presented in Figure~\ref{fig:task_example}. 
This inspires us to the designing criteria that a good data augmentation strategy should uniformly cover these two aspects of coherence.  
Following the criteria, we propose a \textbf{Co}herence evaluation framework via \textbf{U}nified \textbf{D}ata \textbf{A}ugmentation, namely \textbf{CoUDA}, which unifies both global and local aspects of coherence throughout training and inference phase.

\modelname{} involves global and local augmentation to capture the corresponding aspects of coherence.  Regarding global augmentation, we construct negative samples through shuffling, which disrupts the original order of the sentences to induce global incoherence. For local augmentation, our target is to construct negative samples that contain sentences incoherent with the context. While prior rule-based methods, such as swapping a sentence with another from a different text~\citep{shenEvaluatingDocumentCoherence2021}, can also introduce local incoherence, their constructed samples often lack diversity and complexity, potentially failing to capture nuanced aspects of local coherence. To address this, we propose a novel generative augmentation strategy that involves post-pretraining a generative model, and applying two controlling mechanisms to manipulate the difficulty of generated samples. 
By sampling from a generative model, and applying difficulty control, we construct high-quality negative samples to disrupt local coherence.
Finally, in inference phase, 
we design a unified scoring strategy to incorporate both aspects of coherence for overall assessment.

While previous research on coherence evaluation has traditionally adhered to a pairwise ranking setup, we have pioneered a pointwise coherence scoring setting that we believe is more relevant in real-world scenarios. On \textsc{SummEval}\cite{fabbriSummEvalReevaluatingSummarization2021}, our \modelname{} exhibits remarkable improvements in pointwise scoring compared to prior methods, including GPT-4-based metrics. Despite not being specifically tailored for pairwise ranking, our model outperforms previous ranking models on both the \textsc{INSteD-CNN} and \textsc{INSteD-Wiki} datasets~\cite{shenEvaluatingDocumentCoherence2021}. Furthermore, \modelname{} is a lightweight model with only 233M parameters.
To sum up, our contributions are as follows:

\begin{itemize}[leftmargin=*]
    \item We propose \modelname{}, a  data augmentation framework  inspired by  linguistic  theory of discourse structure, which uniformly models both  global and local coherence aspects of a discourse.
    \item We propose a novel generative augmentation strategy, which utilizes the power of the pretrained language model via post-pretraining and two  mechanisms for  sample difficulty control.
    \item Comprehensive experiments  in coherence evaluation show \modelname{} with only 233M  parameters  achieves SOTA performance, even surpassing GPT-3.5 and GPT-4  based metrics.  

\end{itemize}

\section{\modelname{} Framework}

\begin{figure*}
    \centering
    \includegraphics[width=0.8\textwidth]{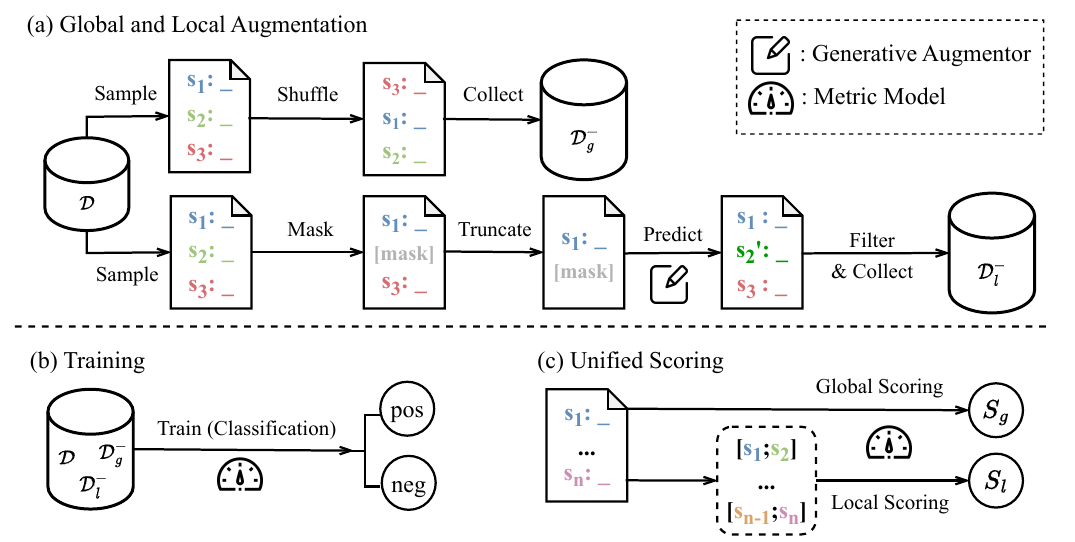}
    \caption{Overview of our proposed \modelname{} framework. (a): First, we use global and local augmentation to create negative samples $\mathcal{D}_g^-$ and $\mathcal{D}_l^-$, respectively. (b): Then, we combine $\mathcal{D}_g^-$ and $\mathcal{D}_l^-$ with the original discourses $\mathcal{D}$ to train our metric model via coherence/incoherence classification. (c): In inference phase, our metric model scores the whole discourse for global score $S_g$, and scores each consecutive sentence pairs for local score $S_l$. $S_g$ and $S_l$ are combined to produce the final coherence score.}
    \label{fig:framework}
\end{figure*}

In this section, we introduce our \modelname{} framework, as illustrated in Figure~\ref{fig:framework}. First, we use global and local augmentation to create negative samples that have relatively poor global coherence and local coherence, respectively. To be specific, we use sentence shuffling for global augmentation, and design a generative strategy for local augmentation. Our generative strategy involves post-pretraining a generative model, and applying two controlling mechanisms to control the difficulty of generated samples. Then we combine the constructed negative samples with the original discourses, which serves as positive samples, to train our metric model for coherence/incoherence classification. In inference phase, we utilize a unified scoring strategy to incorporate global and local coherence for overall assessment.

\subsection{Preliminaries}

\paragraph{Task Formulation.}
Given a discourse that  contains multiple sentences $D=\{s_1, s_2, \dots, s_n\}$, the goal of a coherence evaluator $f_\theta$ is to assess its degree of coherence  by a logit score $f_\theta(D) \in [0, 1]$ (the higher the better).
Ideally, $f_\theta(D)=1$ represents that $D$ is perfectly coherent, while $f_\theta(D)=0$ indicates the opposite. Different from previous work that additionally relies on references ~\citep{zhaoDiscoScoreEvaluatingText2023} or source inputs ~\citep{zhongUnifiedMultiDimensionalEvaluator2022}, we evaluate coherence in this more concise framework that solely takes the discourse as the input.
That is more appropriate for evaluation as  coherence  is an intrinsic quality of a discourse.

\paragraph{Data Augmentation.}  Data augmentation aims to artificially create additional training samples by manipulating existing data.
For a discriminative setting, we need both positive and negative samples for training. 
In terms of coherence evaluation, since a natural discourse $D$ is intrinsically coherent, we focus on applying data augmentation to construct negative samples, i.e. incoherent samples $D^-$. Afterwards, the created incoherent discourses $D^-$  and the original discourse $D$ respectively serve as negative and positive samples to train $f_\theta$.

In the following, we introduce how we construct our two types of negative samples via \textit{global augmentation} and \textit{local augmentation } in details.

\subsection{Global Augmentation}
To construct samples that have relatively poor global coherence, we disrupt the original appropriate organization of sentences in $D$.
Concretely, we shuffle the order of sentences in $D$ to effectively disrupt its global coherence. As
illustrated in Figure~\ref{fig:framework}(a), by shuffling   $D=\{s_1, s_2, s_3\}$, we can construct a negative sample $D_g^-=\{s_3, s_1, s_2\}$.

\subsection{Local Augmentation}

Local augmentation aims to construct samples with  relatively poor local  coherence using the original discourse $D$.
Intuitively, we can realize it  by  replacing a sentence $s_k \in D$ with a substitute $s_k'$ that is incoherent with the leftover discourse $D \backslash s_k$.
This is based on the insight that, through such a replacement, $s_k'$ will decrease  local  coherence of the discourse by introducing an incoherent transition of attention between sentences. 

Subsequently, the most important question is how to find a suitable  $s_k'$ in practice. However,
most prior studies introduce such incoherent elements via heuristic rules, resulting in  $s_k'$ that has \textbf{very weak relevance or even irrelevant} with the remaining discourse $D \backslash s_k$.
For example, INSteD~\citep{shenEvaluatingDocumentCoherence2021} obtains $s_k'$ by extracting sentence of the highest n-gram overlap with $s_k$ from another discourse.
As a result, their introduced local augmentation samples are too easy to train a powerful coherence evaluator.

To construct samples with a higher level of local incoherence, we propose to construct $s_k'$ in a generative way.
Specifically, we train a generative augmentor $G$ to reconstruct $s_k$  based  on $D \backslash s_k$ and use its generated  sentence  $s_k' \sim G(s_k|D \backslash s_k)$ to replace $s_k$.
The strong performance of pretrained generation model will ensure  $s_k'$ to meet the basic standard of fluency and relevance with regard to $D \backslash s_k$.
Meanwhile, due to the intrinsic limitation of autoregressive generation, the reconstructed  $s_k'$ will frequently be incoherent with $D \backslash s_k$, making it possible to construct negative samples in a generative way.
To  further ensure that $s_k'$ conveys the  local incoherence we expect, we design two controlling mechanisms during the  inference  of $G$.
These two mechanisms, \textit{context truncation} and  \textit{coherence filtering }, constraints  $s_k'$ to be neither too strong (perfectly coherent with $D \backslash s_k$)  nor too easy (the incoherence that is too obvious).
Overall, by replacing $s_k$ with $s_k'$, we construct a much stronger negative  sample, which conveys high-level local incoherence while maintaining the basic relevance and fluency with $D \backslash s_k$. 
In the following, we will introduce our \textit{generative augmentor}, \textit{context truncation }and \textit{coherence filtering } in details.

\paragraph{Generative Augmentor.}
 Given discourse $D$, we uniformly sample  $s_k$ from $D$'s non-opening and non-closing sentences.
 Next, we train a text generation model $G$ by learning to reconstruct $s_k$ based on the leftover discourse $D \backslash s_k$.
Following recent popular text generation paradigm, this can be done by selecting $G$ as a  transformer-based sequence-to-sequence model and  maximizing the likelihood  of $G(s|D \backslash s_k)$ autoregressively.

We also notice that Gap Sentences Generation (GSG), the pretraining task of PEGASUS~\citep{zhangPEGASUSPretrainingExtracted2020}, takes the  similar form of 
reconstructing sentences.
But we cannot directly apply PEGASUS as G, because GSG is  specially designed for summarization, which requires predicting multiple   salient sentences in the discourse.
By contrast, our sentence reconstruction task aims to capture the coherence relation between an arbitrary sentence $s$ and the leftover discourse.
In practice, we also find $s_k'$ generated by  PEGASUS often serve as summaries of the leftover discourse, rather than being coherent with it.
Thus, instead of directly applying PEGASUS, we leverage this similarity of tasks and use PEGASUS for initialization.
In this way, we inherit the effectiveness of pretrained model.

After the generative augmentor is trained, we use it to predict $s_k'$ with two controlling mechanisms:

\paragraph{Context Truncation.} 
Due to the strong generation ability of generative augmentor, $s'_k$ may be highly coherent with $D \backslash s_k$, which is not the negative sample we expect.
To ensure  $s'_k$ to convey the local incoherence,  we develop a context truncation mechanism to restrict the model's generation to only partially coherent with the context. 
Specifically, given $D \backslash s_k = \{s_1,s_2, ... ,[mask], ..., s_n\}$ with $s_k$ masked, we randomly choose to truncate the context before or after the mask token, i.e., the input for our generative augmentor is either $\{s_1, s_2, ... , [mask]\}$ or $\{[mask], ..., s_n\}$. Take the former as an example, without information from subsequent text, the model is only able to generate predictions that are coherent with preceding text.

\paragraph{Coherence Filtering.} In addition to context truncation, we also perform coherence filtering to remove negative samples that are too easy. We utilize \textsc{UniEval}~\citep{zhongUnifiedMultiDimensionalEvaluator2022} to score the coherence of each sample and eliminates samples with coherence scores below a filtering threshold $\delta$.

\subsection{Training and Unified Scoring}
\label{sec:metric_model}

\paragraph{Training.} We combine the original discourses with negative samples constructed via global and local augmentation to train our metric model, as illustrated in Figure~\ref{fig:framework}(b). 
We utilize the classification setup, based on findings~\citep{steenHowFindStrong2022} that indicate its superior performance in coherence evaluation and downstream tasks, as opposed to the commonly used pairwise ranking setup. Specifically, we train our metric model to distinguish each sample as coherent or incoherent through binary cross entropy loss. For implementation details, please refer to Appendix~\ref{app:implement}.

\paragraph{Unified Scoring.} For a comprehensive evaluation of discourse coherence, our \modelname{} further includes a unified scoring strategy, as presented in Figure~\ref{fig:framework}(c). Specifically, our model first assigns a score conditioned on the whole discourse to represent its global coherence level:
\begin{equation}
    S_g = f_{\theta}(D)
\end{equation}
Then, since global scoring may fail to effectively capture the fine-grained coherence between sentences, we extract consecutive sentence pairs $[s_i;s_{i+1}]$ from the discourse and have our model evaluate the inter-sentential coherence $S_l^{i}$ of each pairs, where $1\le i \le n-1$:
\begin{equation}
    S_l^{i} = f_{\theta}([s_i;s_{i+1}])
\end{equation}
Notably, although our model is trained for scoring the whole discourse, rather than sentence pairs, the training data includes discourse samples with only two sentences. As a result, our model can generalize to scoring sentence pairs as well. Afterwards, we obtain local coherence score for discourse $D$ by averaging each sentence pair's coherence score:
\begin{equation}
    S_l = \text{Average}(\{S_l^{1},...,S_l^{n-1}\})
\end{equation}
The global and local scores are then combined via interpolation to form the overall coherence score:
\begin{equation}
    \text{Score} = (1-\lambda) \cdot S_g + \lambda \cdot S_l
\end{equation}
where $\lambda \in [0,1]$ controls the weight. This unified design also aligns with the coherence rating process of human readers, who consider both discourse organization as a whole, and smooth transitions of focused items between adjacent sentences.

\section{Experimental Setup}

\subsection{Evaluation Tasks}
We perform meta-evaluation on the proposed metric model in two task settings, i.e. pointwise scoring and pairwise ranking.

\noindent \textbf{Pointwise Scoring} involves assigning coherence scores to text summarization samples and evaluating the correlation between model-assigned scores and human-rated scores. This task closely simulates real-world scenarios. To determine the accuracy of the assigned scores, we compute the correlation coefficients between the model-generated scores and human ratings using \textit{Spearman}~\citep{spearman}, \textit{Pearson}~\citep{pearson}, and \textit{Kendall's tau}~\citep{kendall}. Following previous work, these correlation scores are reported at both \textit{sample-level} and \textit{dataset-level} (See Appendix~\ref{app:implement} for their definitions).

\noindent \textbf{Pairwise ranking} requires the metric models to determine the more coherent option when presented with two candidates. This task serves as an alternative when absolute scores are unavailable, relying solely on relative coherence rankings. For this task, we use accuracy as performance metric.

\subsection{Evaluation Datasets}
For pointwise scoring, we evaluate model performance on \textsc{SummEval}~\citep{fabbriSummEvalReevaluatingSummarization2021}, which is a meta-evaluation benchmark for summarization that contains 100 articles with summaries generated by 16 different systems. For each summary, it offers human annotated scores in terms of fluency, coherence, consistency, and relevance.\footnote{In this paper, we focus on discourse coherence, so we neglect coherence evaluation datasets on dialogue.}

In pairwise ranking, we evaluate model performance on \textsc{INSteD}~\citep{shenEvaluatingDocumentCoherence2021}, which is an intruder sentence detection dataset constructed using discourses from CNN and Wikipedia. We denote these two parts as \textsc{INSteD-CNN} and \textsc{INSteD-Wiki}. In this dataset, incoherent discourses are created by randomly substituting a sentence with another one selected using n-gram overlap from different discourses.

\subsection{Baselines Models}
Though more applicable in real scenarios, few work in coherence evaluation has pioneered in pointwise scoring. For a comprehensive performance comparison, we include  baselines models from three categories: \textbf{1) Pairwise Coherence Evaluators}: \textsc{UNC}~\citep{moonUnifiedNeuralCoherence2019} and \textsc{MultiNeg}~\citep{jwalapuramRethinkingSelfSupervisionObjectives2022}. \textsc{UNC} captures different levels of coherence via a LSTM-based Siamese architecture; \textsc{MultiNeg}\footnote{The original \textsc{MultiNeg} model is backboned with XLNet and trained on the WSJ dataset. For fair comparison, we retrained this model from ALBERT-xxlarge, using the same part of Wikipedia and CNN data. Notably, due to its use of two encoders, \textsc{MultiNeg} has twice the number of parameters compared to \modelname{}.} mines hard negative samples constructed via sentence shuffling to train pairwise coherence ranking models. 2) \textbf{General Evaluators}: \textsc{BartScore}~\citep{yuanBARTScoreEvaluatingGenerated2021}, \textsc{UniEval}~\citep{zhongUnifiedMultiDimensionalEvaluator2022}. \textsc{BartScore} treats text evaluation as a generation task, utilizing \textsc{BART} to assign quality scores for a specific dimension. \textsc{UniEval} reframes text evaluation as a Boolean Question Answering task. Backboned with \textsc{T5}, it is trained with rule-based local augmentation for coherence evaluation. \textbf{3) Large Language Models}: \textsc{G-Eval}~\citep{liuGEvalNLGEvaluation2023} uses LLMs with chain-of-thoughts to assign quality scores. We experiment with two versions using GPT-3.5-Turbo / GPT-4, respectively denoted as \textsc{G-Eval-3.5 / 4}.  We include more details about  using \textsc{UniEval} and \textsc{G-Eval} in Appendix~\ref{sec:app_unieval_source} and ~\ref{sec:app_gpt_template}, respectively.

\begin{table*}[t]
\centering
\footnotesize
\renewcommand{\arraystretch}{1.1}
\setlength\tabcolsep{16pt}
\begin{tabular}{lccccccc}
\toprule
\multirow{2.5}{*}{\textbf{Model}} &
\multirow{2.5}{*}{\textbf{\#Param.} $\downarrow$ } &
\multicolumn{3}{c}{\textbf{Sample-Level} $\uparrow$} &
\multicolumn{3}{c}{\textbf{Dataset-Level} $\uparrow$} \\
\cmidrule(r){3-5} \cmidrule(r){6-8} & & $\rho$ & $r$ & $\tau$ & $\rho$ & $r$ & $\tau$ \\
\midrule
\textsc{UNC} & - & 18.8 & 27.8 & 14.1 & 19.8 & 24.3 & 14.0\\
\textsc{MultiNeg} & 466M & 44.6 & 48.1 & 34.0 & 47.7 & 47.8 & 34.3 \\ \midrule
\textsc{BartScore} & 406M & 44.8 & 45.8 & 34.2 & 40.8 & 43.4 & 29.2 \\
\textsc{UniEval} & 770M & 56.7 & 57.8 & 43.6 & 58.7 & 55.6 & 42.3 \\ \midrule
\textsc{G-Eval-3.5} & >10B & 47.0 & 48.4 & 40.3 & 43.5 & 43.8 & 35.3\\
\textsc{G-Eval-4}\dag{} & >100B & 58.2 & - & 45.7 & - & - & -\\ \midrule
\modelname{} (\textit{ours})& 233M & \textbf{60.0} & \textbf{62.1} & \textbf{46.0} & \textbf{65.6} & \textbf{64.2} & \textbf{47.8} \\
\bottomrule
\end{tabular}
\caption{Sample-level and dataset-level Spearman ($\rho$) / Pearson ($r$) / Kendall ($\tau$) correlations with human ratings on \textsc{SummEval}.  Best results in each column are denoted \textbf{in Bold}. \dag{} denotes results reported in the original paper. With only 233M parameters, \modelname{} largely outperforms previous methods, including GPT-4 based methods.}
\label{tab:exp_correlation}
\end{table*}

\subsection{Details of Synthetic Data}

\paragraph{Data Source.} We obtain positive part of data for our framework by sampling from  CNN~\citep{nallapatiAbstractiveTextSummarization2016} and Wikipedia~\citep{yangWikiQAChallengeDataset2015}. For CNN, we utilize its source documents rather than summaries, because the latter is constructed by combining bullet points, hence lacks coherence. For each source document, we randomly select 2 to 5 leading sentences, enabling our 
metric model to generalize to different lengths. The same length constraint is applied on Wikipedia as well. Concretely, we sample 10,000 documents each from CNN and Wikipedia, hence obtaining 20,000 positive samples. 

\paragraph{Statistics.} For global coherence, we perform permutation on 5,000 positive samples, and acquire 5,000 negative samples for this aspect. For local coherence, we perform gap sentence generation on the remaining 15,000 positive samples using generative augmentor with context truncation. By setting threshold $\delta$ for confidence filtering to 0.5, we obtain 10,889 positive and negative pairs for this aspect. Hence, the final size of our synthetic data (including positive samples) is 31,778. We split it into 30,000 / 1,178 for training and validation. 

\subsection{Implementation Details.} Our metric model utilizes ALBERT~\cite{lanALBERTLiteBERT2020} as the backbone, benefiting from its sentence order prediction task during pretraining to capture information flow between sentences. Specifically, we use ALBERT-xxlarge with a total of 233M parameters. We set batch size to 32 and learning rate to $1e^{-5}$. Convergence is reached within 3,000 steps. We use the best performing checkpoint on the validation part of synthetic data. Details about generative augmentor are presented in Appendix~\ref{app:implement}. In terms of hyperparameters $\lambda$ and $\delta$, we simply set both of them to $0.5$.

\section{Results}

\begin{table}[t]
  \centering
  \footnotesize
  \renewcommand{\arraystretch}{1.1}
  \setlength\tabcolsep{20pt}
  \begin{tabular}{lcc}
    \toprule
    \textbf{Model} & \textbf{CNN} & \textbf{Wiki} \\
    \midrule
    \textsc{UNC} & 96.4 & 60.5 \\
    \textsc{MultiNeg} & 94.2 & 72.1 \\ \midrule
    \textsc{BartScore} & 70.7 & 58.8 \\
    \textsc{UniEval} & 92.0 & 77.3 \\ \midrule
    \textsc{G-Eval-3.5} & 82.2 & 58.5 \\ \midrule
    \modelname{} & \textbf{98.5} & \textbf{79.1} \\
    \bottomrule
  \end{tabular}
  \caption{Pairwise ranking accuracy on the CNN and Wikipedia split of \textsc{INSteD}.}
  \label{tab:exp_pairwise}
\end{table}

In this section, we show that \modelname{} framework achieves impressive coherence evaluation results on pointwise scoring and pairwise ranking tasks, even when compared with GPT-4 based models. We report average scores across 3 runs with different random seeds.

\subsection{Results on \textsc{SummEval}}

Table~\ref{tab:exp_correlation} presents the sample-level and dataset-level correlations of each model with human ratings on \textsc{SummEval}. Since \textsc{UNC} and \textsc{MultiNeg} are trained through pairwise ranking, their performance on for pointwise scoring is relatively limited. \textsc{BartScore} and \textsc{UniEval} are general evaluators for multiple dimensions such as informativeness and coherence. The former lacks specific training for these dimensions, leading to lower performance, while the latter gain significant improvement through tailored training for coherence. However, \textsc{UniEval} still relies on heuristic rules for augmentation, resulting in limited improvements. The third block presents the results of \textsc{G-Eval-3.5} and \textsc{G-Eval-4}, built upon \textsc{GPT-3.5-Turbo} and \textsc{GPT-4}, respectively. Since there are no exact description of how many parameters GPT-3.5/4 takes, we estimate them as >10B and >100B.

Among baselines models, \textsc{G-Eval-4} achieves highest correlation with human ratings, followed by \textsc{UniEval}, which demonstrates strong performance, even surpassing \textsc{G-Eval-3.5}. Compared with \textsc{UniEval}, \modelname{} consistently shows its superiority on both sample-level correlation~(+3.3/+4.3/+2.4 in $\rho, r, \tau$) and dataset-level correlation~(+6.9/+8.7/+5.4 in $\rho, r, \tau$). 
With only 233M parameters, it also surpasses \textsc{G-Eval-4} in both sample-level Spearman and Kendall correlations by 1.8 and 0.3 points, respectively. This remarkable improvement consolidates the efficacy of our designing criteria.
Additionally, we notice that performance gain in dataset-level correlation is much greater than that of sample-level.

\subsection{Results on \textsc{INSteD}}

Table~\ref{tab:exp_pairwise} presents each model's pairwise ranking accuracy on \textsc{INSteD-Wiki} and \textsc{INSteD-CNN}. Both \textsc{MultiNeg} and \textsc{UNC} achieves impressive accuracy. We suppose it is because they are exactly trained using the pairwise ranking setup.
\textsc{UniEval} also achieves competitive results, which means that specialized training for coherence greatly enhances model performance. Surprisingly, \textsc{G-Eval-3.5} obtains merely above chance accuracy on \textsc{INSteD-Wiki}, indicating that current LLMs are unreliable in pairwise ranking tasks, necessitating further investigation and attention from researchers. Our \modelname{}, though not directly trained under pairwise ranking settings, achieves best results on both \textsc{INSteD-CNN} and \textsc{INSteD-Wiki}, with a performance gain of 2.2 and 1.8 points, respectively.

\begin{table}[t]
  \centering
  \footnotesize
  \renewcommand{\arraystretch}{1.1}
  \setlength\tabcolsep{5pt}
  % \begin{tabular}{ccc|cccccc}
  \begin{tabular}{p{0.4cm}| p{0.4cm} p{0.4cm} | ccc ccc}
    \toprule
    \multirow{2.5}{*}{$G$} & \multirow{2.5}{*}{$L_G$} & \multirow{2.5}{*}{$L_R$} & 
    \multicolumn{3}{c}{\textbf{Sample-Level}} &
    \multicolumn{3}{c}{\textbf{Dataset-Level}} \\
    \cmidrule(r){4-6} \cmidrule(r){7-9} & & & $\rho$ & $r$ & $\tau$ & $\rho$ & $r$ & $\tau$ \\
    \midrule
    \checkmark & & & 56.3 & 57.2 & 43.1 & 58.2 & 56.2 & 42.1 \\
    & \checkmark & & \underline{57.6} & \underline{59.4} & \underline{44.1} & \underline{62.9} & \underline{61.6} & \underline{45.6}  \\
    &  & \checkmark & 53.8 & 56.4 & 41.1 & 59.4 & 60.1 & 43.2   \\ \midrule
    \checkmark & & \checkmark & 56.6 & 59.3 & 42.5 & 61.5 & 61.3 & 44.1  \\
    \checkmark & \checkmark & & \textbf{60.0} & \textbf{62.1} & \textbf{46.0} & \textbf{65.6} & \textbf{64.2} & \textbf{47.8}  \\
    
    \bottomrule
  \end{tabular}
  \caption{Comparison of global augmentation $G$, our generative local augmentation $L_G$, previous rule-based local augmentation $L_R$, and their combinations.}
  \label{tab:ana_data}
\end{table}

\section{Comparison of Augmentation Methods}
\label{sec:data_aug_comp}

In this section, we validate the advantage of our unified data augmentation strategy for coherence scoring over previous data augmentation strategies.

\paragraph{Compared Data Augmentation Methods.} Coherence evaluation emphasizes the sentence structure and organization of a discourse. Due to this special focus, data augmentation strategies designed for other tasks, e.g. EDA~\citep{weiEDAEasyData2019a}, are not directly applicable. Instead, we compare following data augmentation strategies for generating negative samples: \textbf{1) G:} Global augmentation via sentence shuffling~\citep{barzilayModelingLocalCoherence2008}, which is also adopted in our framework. \textbf{2) $\mathbf{L_R}$:} Rule-based local augmentation through sentence intrusion, which employs n-gram overlap to select locally incoherent samples~\citep{shenEvaluatingDocumentCoherence2021}. \textbf{3) $\mathbf{L_G}$:} Our generative local augmentation strategy. \textbf{4) $\mathbf{G+L_R}$ or $\mathbf{G+L_G}$:} Combination of global and local augmentation methods.

\paragraph{Global vs. Local vs. Unified.} In Table~\ref{tab:ana_data}, we can see that unifying global and local augmentation data yields the best human correlation, better than using global or local augmentation alone. This aligns well with the linguistic theory of discourse structure that the organization of discourse segments (global coherence), and the transition of attention or focused items (local coherence) are two key factors of discourse coherence, from which our unified data augmentation framework are inspired.

\paragraph{Generative vs. Rule-based.} Further, we compare the result of generative augmentation vs. rule-based augmentation for modeling local coherence. First, metric model trained with $L_G$ outperforms that of $L_R$ by a large margin on both sample-level correlation (+3.8/+3.0/+3.0 in $\rho, r, \tau$) and dataset-level correlation (+3.5/+1.5/+2.4 in $\rho, r, \tau$). Second, when combined with global augmentation, $G+L_G$ yields significantly superior performance than $G+L_R$. Based on these two aspects, we can conclude that our generative strategy is more effective than rule-based methods.

\section{Analysis}

\label{sec:ana}

\begin{table}[t]
\centering
\footnotesize
\renewcommand{\arraystretch}{1.1}
\setlength\tabcolsep{3pt}
\begin{tabular}{lcccccc}
\toprule
\multirow{2.5}{*}{\textbf{Method}} &
\multicolumn{3}{c}{\textbf{Sample-Level}} &
\multicolumn{3}{c}{\textbf{Dataset-Level}} \\
\cmidrule(r){2-4} \cmidrule(r){5-7} & $\rho$ & $r$ & $\tau$ & $\rho$ & $r$ & $\tau$ \\
\midrule
Generative Augmentor & 30.2 & 33.5 & 22.5 & 26.7 & 25.4 & 18.6 \\ \midrule
+ C.T. & 51.4 & 51.9 & 38.9 & 53.8 & 51.9 & 38.2 \\
+ C.T. + filter $\delta=0.2$ & 53.7 & 51.3 & 41.5 & 55.9 & 47.8 & 40.3 \\
+ C.T. + filter $\delta=0.4$ & 52.8 & 54.0 & 40.4 & 56.8 & 53.7 & 40.1 \\
+ C.T. + filter $\delta=0.6$ & \textbf{55.7} & \textbf{55.3} & \textbf{42.8} & \textbf{58.0} & \textbf{55.9} & \textbf{42.0} \\
+ C.T. + filter $\delta=0.8$ & 46.3 & 42.7 & 35.9 & 49.5 & 41.5 & 35.7 \\
\bottomrule
\end{tabular}
\caption{Analysis of our controlling mechanisms for local augmentation. \textit{C.T.} stands for context truncation. $\delta$ is the threshold for confidence filtering.}
\label{tab:ana_difficult}
\end{table}

\begin{figure}[t]
    \centering
    \includegraphics[width=\linewidth]{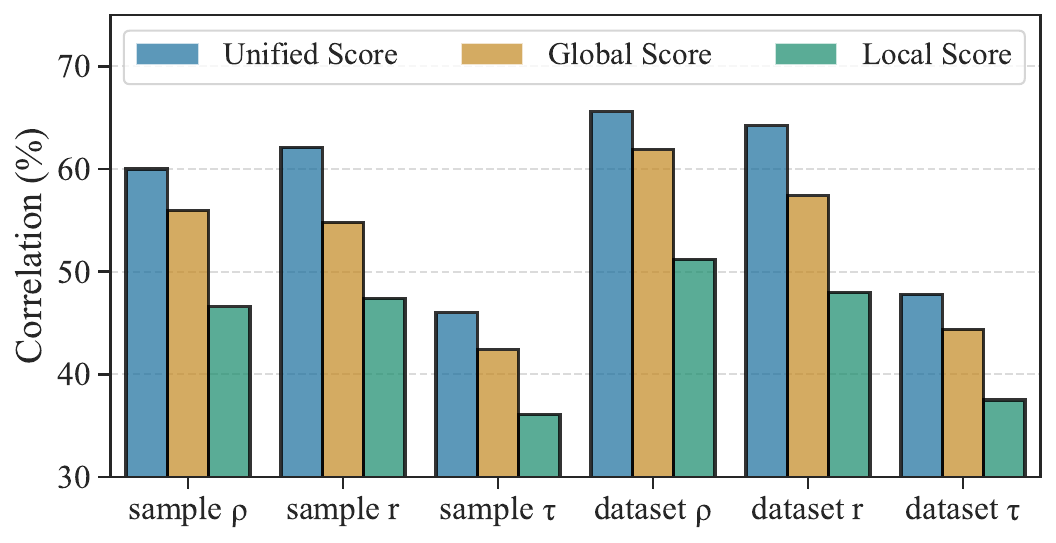}
    \caption{Ablation study of global and local scores in our unified scoring strategy.}
    \label{fig:ana_score}
\end{figure}

\paragraph{Unified Scoring.} First, we study the effectiveness of our unified scoring strategy. Experiment results are demonstrated in Figure~\ref{fig:ana_score}. First, both global and local scores are beneficial in improving human correlation. Additionally, global scores correlate better with human ratings than local scores.

\paragraph{Controlling Mechanisms.} We then analyze the effect of our difficulty controlling mechanisms in local augmentation.
Specifically, we train our metric model separately on local augmentation data constructed under different settings to compare their impacts. Table~\ref{tab:ana_difficult} presents the results. First, we can see that context truncation contributes a significant portion of performance, without which our generative augmentor suffers a severe performance drop of more than 20 points. This demonstrates the effectiveness of constructing partially coherent samples. Second, we find that our confidence filtering mechanism, through which we filter out easy negative samples, also helps model performance. We found that 0.6 is an optimal threshold that can filter out easy examples while ensuring enough amount of training data. We have also provided a case study in Appendix~\ref{app:case_study}.

\begin{figure}[t]
    \centering
    \includegraphics[width=\linewidth]{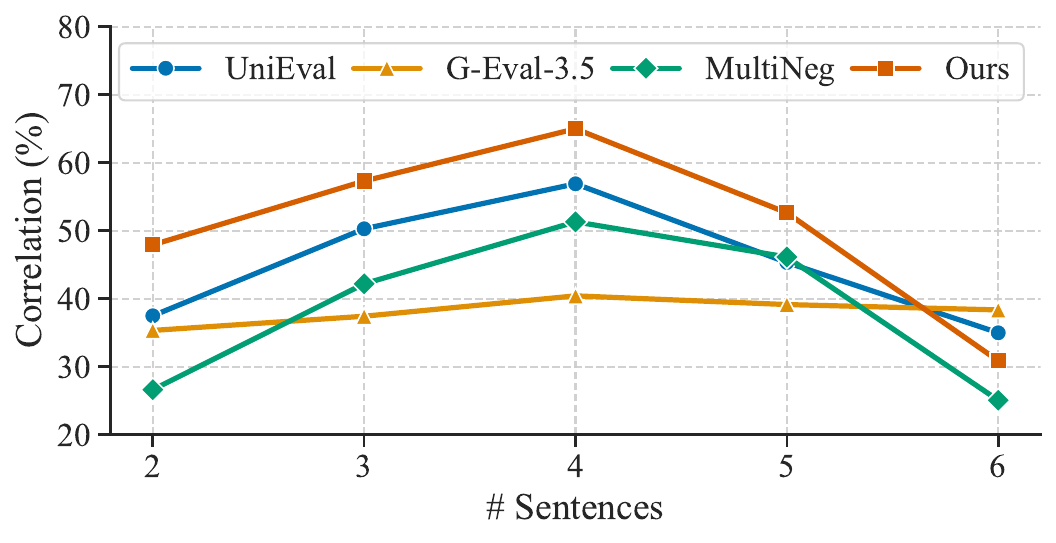}
    \caption{Average of dataset-level Spearman / Pearson / Kendall correlation on \textsc{SummEval} w.r.t. discourses containing different numbers of sentences.
    }
    \label{fig:ana_length}
\end{figure}

\paragraph{Discourse Length.} We compare our model's performance with strong baselines (\textsc{MultiNeg}, \textsc{MultiNeg}, \textsc{G-Eval-3.5}) w.r.t. different discourse length. Concretely, we categorize all 1,600 system summaries of \textsc{SummEval} into different groups according to the sentence numbers they have. We calculate the average of dataset-level Spearman / Pearson / Kendall correlation as defined in Equation~\ref{eqn:dataset_corr} for each group. Figure~\ref{fig:ana_length} presents the results. On average, our model achieves best results when target discourse contains no more than 5 sentences. As the discourse length increases, all models suffer from performance drop, with \textsc{G-Eval-3.5} being the only exception, which renders very steady correlation against length variance. Since each training sample we construct contains no more than 5 sentences~(see Appendix~\ref{app:implement}), we assume \modelname{}'s performance drop can be alleviated by training on samples with more sentences.

\section{Related Work}
\subsection{Coherence Evaluation}

Coherence evaluation measures the organization and structure of a discourse. Due to the paucity of human-annotated training data, previous work has mainly focused on two synthetic tasks: permutation detection and sentence intrusion detection. Permutation detection task~\cite{barzilayModelingLocalCoherence2005,elsnerUnifiedLocalGlobal2007,barzilayModelingLocalCoherence2008, liNeuralNetModels2017} requires the model to distinguish original discourse from its sentence shuffled version. Sentence intrusion detection task~\cite{shenEvaluatingDocumentCoherence2021} determines whether a discourse contains an intruder sentence from another discourse. 

A series of methods have been proposed for these synthetic tasks. \citet{barzilayModelingLocalCoherence2005,barzilayModelingLocalCoherence2008} introduced the popular entity-based model using Centering Theory~\cite{groszCenteringFrameworkModeling1995}. It was further extended to combine with entity-specific features~\cite{elsnerExtendingEntityGrid2011}, convolutional neural networks~\cite{tiennguyenNeuralLocalCoherence2017}, and graph neural networks~\cite{mesgarNeuralGraphbasedLocal2021}. \citet{jwalapuramRethinkingSelfSupervisionObjectives2022} attempted to improve model generalization by training their model purely through self-supervision, with negative samples mined from permutation space. Instead, we propose to improve evaluation performance by unifying different aspects of discourse coherence, as inspired by linguistic theory of discourse structure~\citep{groszAttentionIntentionsStructure1986}. UNC~\citep{moonUnifiedNeuralCoherence2019} captured different levels of coherence via a Siamese architecture that involved bi-linear projection and lightweight convolution-pooling. By contrast, we address this from the perspective of data augmentation rather than model architecture.

\subsection{General Evaluators}

We denote evaluators capable of assessing multiple quality dimensions by altering input and output contents~\cite{yuanBARTScoreEvaluatingGenerated2021}, or adopting different formulas~\cite{scialomQuestEvalSummarizationAsks2021, zhongUnifiedMultiDimensionalEvaluator2022} as general evaluators. A leading trend is to utilize generation model for quality assessment, such as \textsc{BartScore}~\cite{yuanBARTScoreEvaluatingGenerated2021}, \textsc{UniEval}~\cite{zhongUnifiedMultiDimensionalEvaluator2022}.
Apart from that, \textsc{DiscoScore}~\cite{zhaoDiscoScoreEvaluatingText2023} compared the focus matrix between the candidate and the reference to calculate the overall quality score.

With the rise of large language models (LLMs), there has been a growing tendency to use LLMs for evaluation purpose~\citep{wangChatGPTGoodNLG2023,fuGPTScoreEvaluateYou2023,wangLargeLanguageModels2023,liuGEvalNLGEvaluation2023}. \citet{wangChatGPTGoodNLG2023} adopted ChatGPT for NLG evaluation and achieved competitive results in terms of correlation with human judgments. \citet{liuEvaluatingTextCoherence2020} used LLMs with chain-of-thought and a form-filling paradigm to assess the quality of text.

\section{Conclusion}

We propose a unified data augmentation framework called \modelname{}, with the designing criteria to unify both global and local aspects of coherence, as inspired by linguistic theory of discourse structure. This data framework includes global and local augmentation, a classification paradigm for training and a unified scoring strategy for inference. We specifically propose a novel generative augmentation strategy, which involves post-pretraining a generative model, and applying two controlling mechanisms to control the difficulty of generated samples. With only 233M parameters, our framework achieves remarkable improvement over previous methods, including GPT-4 based metrics.

\section*{Limitations}

Our work is still limited in some aspects, particularly in handling extra long discourses. Note that our framework requires assigning coherence scores to all adjacent sentence pairs. While this approach allows for detailed modeling of local coherence between sentences, it may be slow when dealing with documents that contain a large number of sentences.

\section*{Ethics Statement}

Our work complies with the ACL Ethics Policy.
Since all datasets we used are publicly available, we have not identified any significant ethical considerations associated with our work.

% Scientific work published at EMNLP 2023 must comply with the \href{https://www.aclweb.org/portal/content/acl-code-ethics}{ACL Ethics Policy}. We encourage all authors to include an explicit ethics statement on the broader impact of the work, or other ethical considerations after the conclusion but before the references. The ethics statement will not count toward the page limit (8 pages for long, 4 pages for short papers).

\section*{Acknowledgements}
We thank the anonymous reviewers for their helpful comments on this paper. This work was partially supported by 
National Key R\&D Program of China (No. 2022YFC3600402).
%The corresponding author of this paper is Sujian Li.

\bibliography{bib/anthology,bib/custom}
\bibliographystyle{bib/acl_natbib}

\appendix

\section{Details of Data, Generative Augmentor, and Correlation Calculation}
\label{app:implement}

\paragraph{Details of Generative Augmentor.} Our generative augmentor is initialized with PEGASUS-Large using the checkpoint in Huggingface. We train it on the positive samples mentioned above, with batch size set to 32. Convergence is reached within 5,000 steps. To avoid data leakage in training and prediction, we split our positive samples into part A and part B, each with 20,000 samples. We first train our model solely  on part A, and use it to construct negative samples for part B. Then we train a new model solely on part B, and use it to construct negative samples for part A.

\paragraph{Sample-Level and Dataset-Level Correlation.} Suppose we have $n$ documents in a dataset, for each document $d_i, i\in\{1,...,n\}$, we have $J$ system outputs. Let $o_{ij}, j\in\{1,...,J\}$ be the output of the $j^{th}$ system for the $i^{th}$ document, $K$ be a correlation measure, $f_\theta$ and $f_h$ be metric model and human evaluation, respectively, sample-level correlation and dataset-level correlations can be calculated as follows:

\noindent (1) \textit{Sample-level} correlation.
\begin{equation}
\begin{aligned}
K^{sample}=& \frac{1}{n}\sum_{i=1}^{n}K([f_\theta(o_{i1}),...,f_\theta(o_{iJ})] , \\ & [f_h(o_{i1}),...,f_h(o_{iJ})])
\end{aligned}
\end{equation}

\noindent (2) \textit{Dataset-level} correlation.
\begin{equation}
\begin{aligned}
K^{dataset}= & K([f_\theta(o_{11}),...,f_\theta(o_{nJ})], \\& [f_h(o_{11}),...,f_h(o_{nJ})])
\end{aligned}
\label{eqn:dataset_corr}
\end{equation}

\section{Case Study}
\label{app:case_study}
We demonstrate an example of substitute sentences selected or generated using different methods in Table~\ref{tab:local_aug_example}. \textsc{Rule} selects the substitute sentence through n-gram overlap, resulting in a relatively easy samples, as the selected sentence is very incoherent with the context. PEGASUS generates a sentence that summarizes the remainder, rather than being coherent with the context. The prediction of our generative augmentor is highly coherent with the context, making it difficult to be distinguished as negative. Through context truncation, we obtain a partially coherent prediction, which is only coherent with proceeding sentences.

\begin{table}[t]
    \small
    \centering
    \begin{tabularx}{\linewidth}{X}
    \toprule
     \textit{Context} \\ \midrule
     The cities of Annecy, Munich and Pyeongchang will battle it out to host the 2018 Winter Olympics. \textbf{[mask]} The International Olympic Committee have confirmed they have received applications from France, Germany and South Korea ahead of this week's deadline. \\ \midrule
     \textit{Predictions} \\ \midrule
     \textbf{\textsc{Rule}:} Thousands of South Koreans gathered at the foot of a ski jump well past midnight in a passionate display of excitement that included fireworks,  singing,  dancing,  picnicking and kimchi -- the traditional Korean side dish. \\ \midrule
     \textbf{PEGASUS:} The cities of Annecy, Munich and Pyeong-chang will battle it out to host the 2018 Winter Olympics. \\ \midrule
     \textbf{Generative Augmentor (GA):} The French resort of Annecy, the German city of Munich and the South Korean city of Pyeongchang have all submitted bids to host. \\ \midrule
     \textbf{GA w/ Context Truncation:} The International Olympic Committee's Executive Board will meet on Wednesday in Copenhagen to pick the host. \\

     \bottomrule

    \end{tabularx}
    \caption{Comparison of different local augmentation strategies.}
    \label{tab:local_aug_example}
\end{table}

\section{Performance Comparison of UniEval w/ or w/o Source Document}

\label{sec:app_unieval_source}

\begin{table}[t]
    \centering
    \footnotesize
    \renewcommand{\arraystretch}{1.1}
    \setlength\tabcolsep{5pt}
    % \begin{tabular}{ccc|cccccc}
    \begin{tabular}{l|ccc ccc}
      \toprule
      \multirow{2.5}{*}{Src Doc.} & 
      \multicolumn{3}{c}{\textbf{Sample-Level}} &
      \multicolumn{3}{c}{\textbf{Dataset-Level}} \\
      \cmidrule(r){2-4} \cmidrule(r){5-7} & $\rho$ & $r$ & $\tau$ & $\rho$ & $r$ & $\tau$ \\
      \midrule
      \multicolumn{7}{c}{\textit{UniEval}} \\ \midrule
      Empty ("") & 56.7 & 57.8 & 43.6 & 58.7 & 55.6 & 42.3 \\
      Original Src & 57.5 & 55.4 & 44.2 & 59.2 & 53.3 & 42.5 \\ \midrule
      \multicolumn{7}{c}{\textit{CoUDA}} \\ \midrule
      None & 60.0 & 62.1 & 46.0 & 65.6 & 64.2 & 47.8 \\
      \bottomrule
    \end{tabular}
    \caption{Performance Comparison of UniEval w/ or w/o Source Document.}
    \label{tab:app_source}
\end{table}

Recall that \textsc{UniEval} requires a source document as input when assessing coherence. Since our framework solely takes the discourse as input, we set its source document to empty string for fair comparison. In this section, we conduct additional experiments to explore how the source document influences coherence evaluation for \textsc{UniEval}. Results are presented in Table~\ref{tab:app_source}. It can be observed that whether the source is provided does not have a significant impact on the performance of \textsc{UniEval}. This further consolidates our assumption that coherence is an intrinsic quality of discourse that its evaluation does not require other inputs. Furthermore, even with original source provided to \textsc{UniEval}, \modelname{}'s performance remains substantially superior, verifying the effectiveness of our proposed method.

\section{Skewed Template to use G-Eval for Pairwise Ranking}
\label{sec:app_gpt_template}
\begin{figure}[t]
    \centering
    \includegraphics[width=\linewidth]{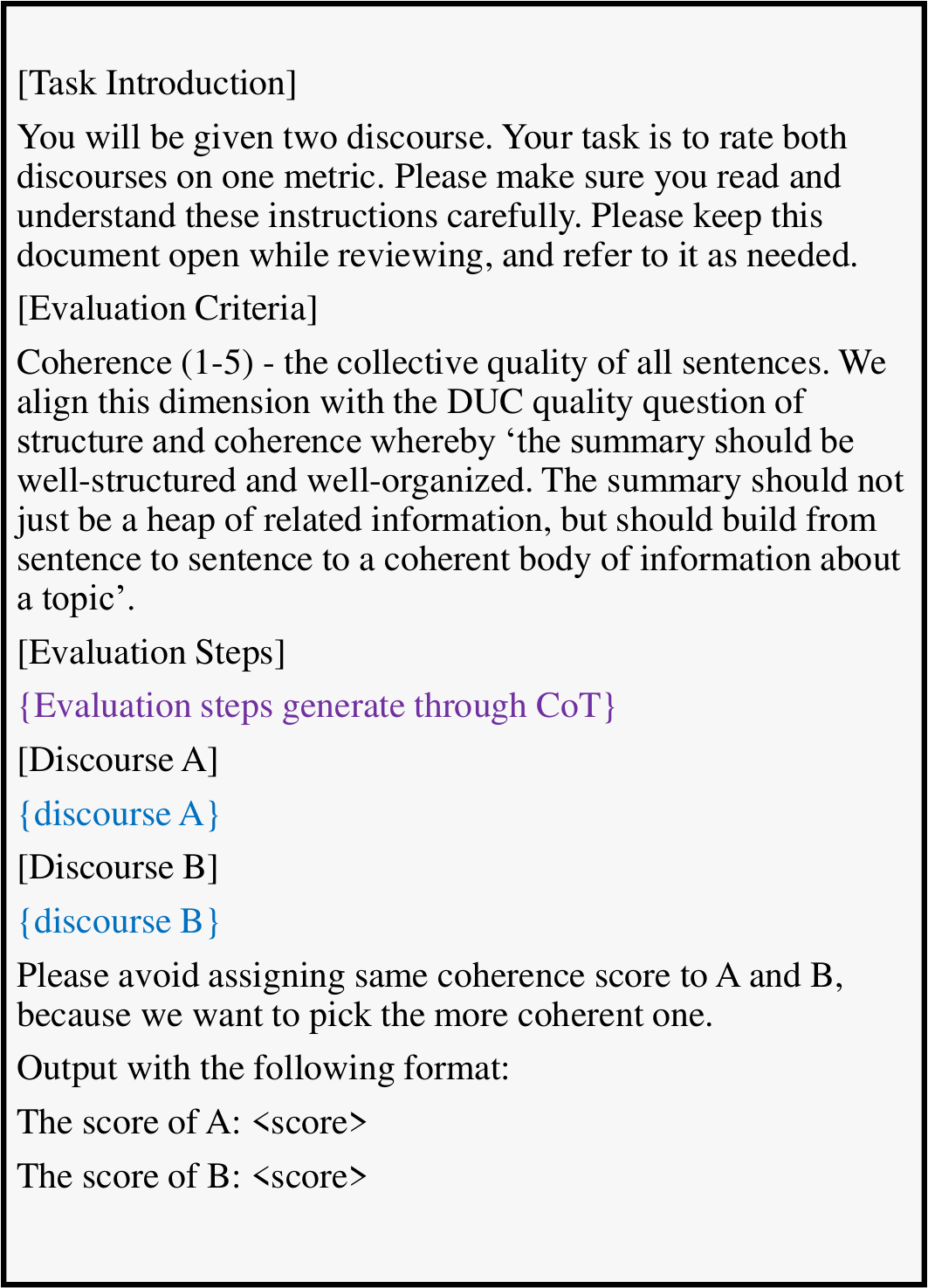}
    \caption{Skewed Template for G-Eval-3.5 in pairwise ranking. We adopt the Balanced Position Calibration strategy proposed by \citet{wangLargeLanguageModels2023} to alleviate positional bias of LLMs}
    \label{fig:app_template}
\end{figure}

Skewed template to use G-Eval for pairwise ranking is presented in Figure~\ref{fig:app_template}. We adopt the Balanced Position Calibration strategy proposed by \citet{wangLargeLanguageModels2023} to alleviate positional bias of LLMs.

\section{The choice of Weight Parameter $\lambda$}

\begin{figure}
    \centering
    \includegraphics[width=\linewidth]{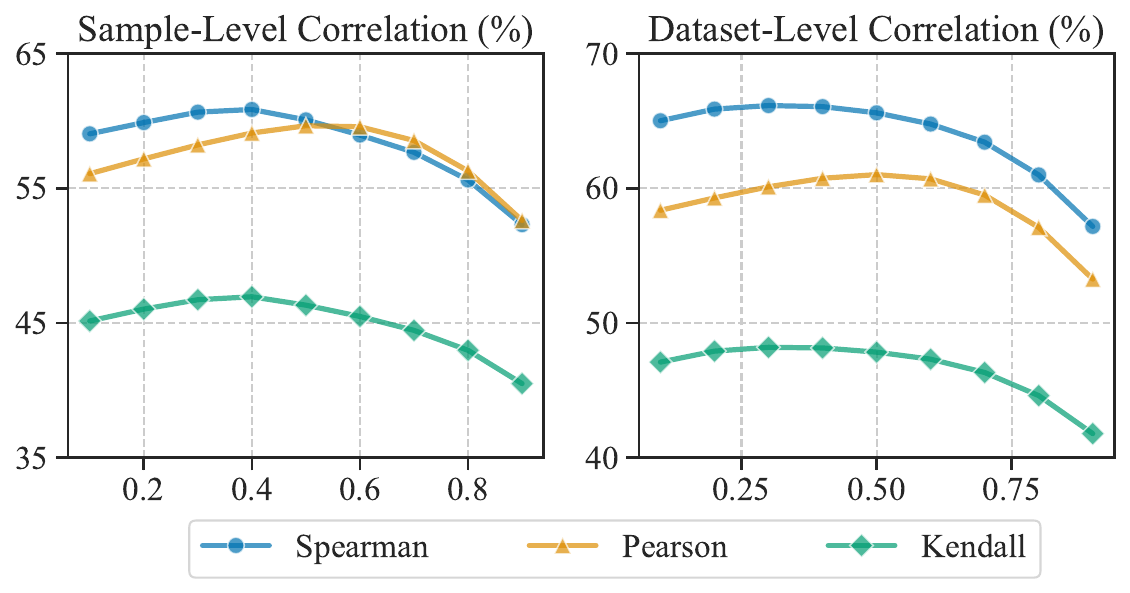}
    \caption{Sample-level correlation and dataset-level correlation on \textsc{SummEval} with different weight parameter $\lambda \in [0,1]$ for global and local coherence score.
    }
    \label{fig:ana_ratio}
\end{figure}

Figure~\ref{fig:ana_ratio} shows the results of varying weight parameter $\lambda$ for global and local coherence score. We see that the best weight for Spearman correlation and Kendall correlation is around 0.4, while the best weight for Pearson correlation is around 0.6. 

\end{document}